\documentclass[10pt,twocolumn,letterpaper]{article}

\usepackage{cvpr}
\usepackage{times}
\usepackage{epsfig}
\usepackage{graphicx}
\usepackage{amsmath}
\usepackage{amssymb}
\usepackage{authblk}

\usepackage{subfigure}
\usepackage{tabularx}
\usepackage{booktabs}
\usepackage{array}
\usepackage{bigstrut}
\usepackage{setspace}
\usepackage{ragged2e}
\usepackage{url}

\usepackage{fancyhdr}		

\newcolumntype{P}[1]{>{\RaggedRight\hspace{0pt}}p{#1}}
\newcolumntype{C}[1]{>{\centering}m{#1}}

\cvprfinalcopy 

\def\cvprPaperID{****} 

\begin{document}

\title{Cultural Event Recognition with Visual ConvNets and Temporal Models}


\author[*]{Amaia Salvador}
\author[**]{Matthias Zeppelzauer}
\author[*]{Daniel Manch{\'o}n-Vizuete}
\author[*]{Andrea Calafell}
\author[*]{Xavier Gir{\'o}-i-Nieto}

\affil[*]{Universitat Politecnica de Catalunya (UPC)\\ Barcelona, Catalonia/Spain\\
amaia.salvador@upc.edu} 
\affil[**]{St. Poelten University of Applied Sciences\\ St. Poelten, Austria\\ matthias.zeppelzauer@fhstp.ac.at}

\maketitle

\begin{abstract}
%
%

This paper presents our contribution to the ChaLearn Challenge 2015 on Cultural Event Classification. The challenge in this task is to automatically classify images from 50 different cultural events. Our solution is based on the combination of visual features extracted from convolutional neural networks with temporal information using a hierarchical classifier scheme. We extract visual features from the last three fully connected layers of both CaffeNet (pretrained with ImageNet) and our fine tuned version for the ChaLearn challenge. We propose a late fusion strategy that trains a separate low-level SVM on each of the extracted neural codes. The class predictions of the low-level SVMs form the input to a higher level SVM, which gives the final event scores. We achieve our best result by adding a temporal refinement step into our classification scheme, which is applied directly to the output of each low-level SVM. Our approach penalizes high classification scores based on visual features when their time stamp does not match well an event-specific temporal distribution learned from the training and validation data. Our system achieved the second best result in the  ChaLearn Challenge 2015 on Cultural Event Classification with a mean average precision of 0.767 on the test set.

\end{abstract}
\section{Motivation}
\label{sec:motivation}

Cultural heritage is broadly considered a value to be preserved through generations.
From small town museums to worldwide organizations like UNESCO, all of them aim at keeping, studying and promoting the value of culture.
Their professionals are traditionally interested in accessing large amounts of multimedia data in rich queries which can benefit from image processing techniques.
For example,  one of the first visual search engines ever, IBM's QBIC \cite{flickner1995query}, was showcased for painting retrieval from the Hermitage Museum in Saint Petersburg (Russia).

\begin{figure}%
		\includegraphics[width=\linewidth]{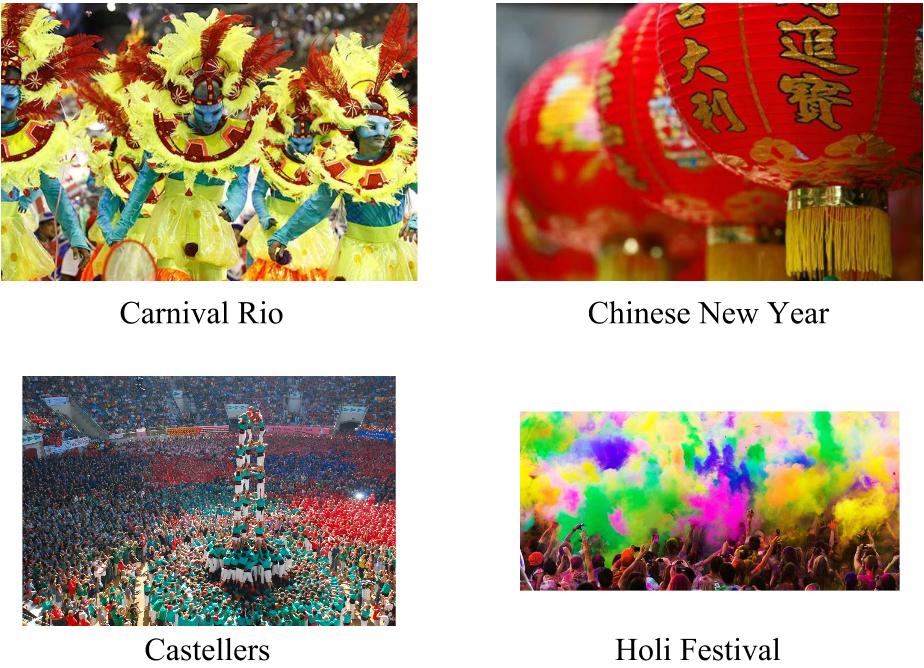}
		\caption{Examples of images depicting cultural events.}
		\label{fig:examples}
\end{figure}

A cultural expression which is typically not found in a museum are social events.
Every society has created through years collective cultural events celebrated with certain temporal periodicity, commonly yearly.
These festivities may widely spread geographically, like the Chinese New Year's or Indian Holi Festival, or much more localized like the Carnival in Rio de Janeiro or the \textit{Castellers} (human towers) in Catalonia.
An image example for each of these four cultural events is presented in Figure \ref{fig:examples}.
All of them have a deep cultural and identity nature that motivates a large amount of people to repeat very particular behavioral patterns.

%
%

The study and promotion of such events has also benefited from the technological advances that have popularized the acquisition, storage and distribution of large amounts of multimedia data.
Cultural events across the globe are at the tip of a click, improving both the access of culture lovers  to rich visual documents, but also their touristic power or even exportation to new geographical areas.

However, as in any classic multimedia retrieval problem, while the acquisition and storage of visual content is a popular practice among event attendees, their proper annotation is not.
While both personal collections and public repositories contain a growing amount of visual data about cultural events, most of it is not easily available due to the almost non-existent semantic metadata.
Only a minority of photo and video uploaders will add the simplest form of annotation, a tag or a title, while most users will just store their visual content with no further processing.
Current solutions will mostly rely in on temporal and geolocation metadata attached by the capture devices, but also these sources are unreliable for different reasons, such as erroneous set up of the internal clock of the cameras, or the metadata removal policy applied in many photo sharing sites to guarantee privacy.

Cultural event recognition is a challenging retrieval task because of its strong semantic dimension. The goal of cultural event recognition is not only to find images with similar content, but further to find images that are semantically related to a particular type of event. Images of the same cultural event may also be visually  different. Thus, major research questions in this context are, (i) if content-based features are able to represent the cultural dimension of an event and (ii) if robust visual models for cultural events can be learned from a given set of images.

\begin{figure*}%
		\includegraphics[width=\textwidth]{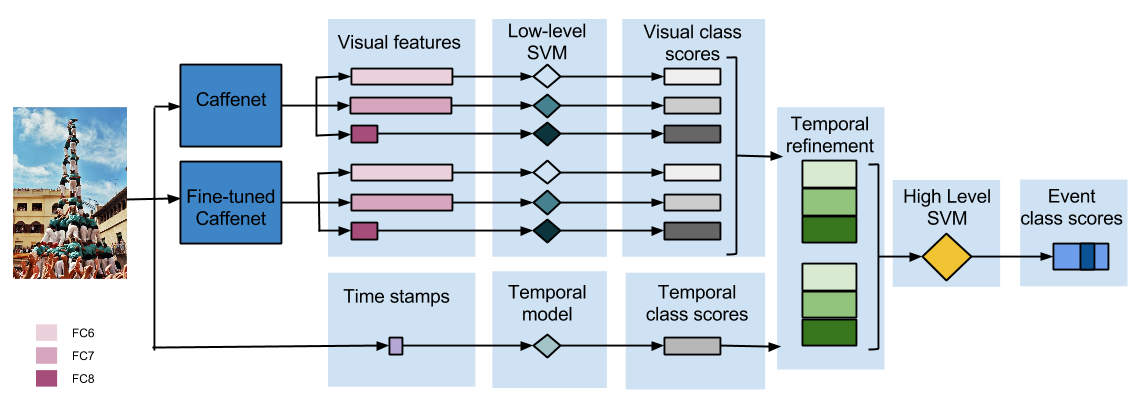}
		\caption{Global architecture of the proposed system.}
		\label{fig:pipeline}
\end{figure*}

In our work, we addressed the cultural event recognition problem in photos by combining the visual features extracted from convolutional neural networks (convnets) with metadata (time stamps) of the photos in the hierarchical fusion scheme shown in Figure \ref{fig:pipeline}. The main contributions of our paper are:

\begin{itemize}
\item Late fusion of the neural codes from both the fine-tuned and non-fine-tuned fully connected layers of the CaffeNet \cite{jia2014caffe} convnet.
\item Generation of spline-based temporal models for cultural events based on photo metadata crawled from the web. 
\item Temporal event modeling to refine visual-based classification as well as noisy data augmentation. 
\end{itemize}

This paper is structured as follows. Section \ref{sec:related_work} overviews the related work, especially in the field of social event detection and classification.
Section \ref{sec:temporal} describes a temporal modeling of the cultural events which has been applied both on the image classification and data augmentation strategies presented in Section \ref{sec:classification} and Section \ref{sec:DataAugmentation}, respectively. 
Experiments on the ChaLearn Cultural Event Dataset \cite{escalera2015chalearn} are reported in Section \ref{sec:experiments} and conclusions drawn in Section \ref{sec:conclusions}.

This work was awarded with the 2nd prize in the ChaLearn Challenge 2015 on Cultural Event Classification.
Our source code, features and models are publicly available online\footnote{\url{https://imatge.upc.edu/web/resources/cultural-event-recognition-computer-vision-software}}.
\section{Related work}
\label{sec:related_work}

The automatic event recognition on photo and video collections has been broadly addressed from a multimedia perspective, further than just the visual one.
Typically, visual content is accompanied by descriptive metadata such as a time stamp from the camera or an uploading site, a geolocation from a GPS receiver or some text in terms of a tag, a title or description is available.
This additional contextual data for a photo is highly informative to recognize the depicted semantics.


Previous work on social events has shown that temporal information provides strong clues for event clustering \cite{zaharievacross15}. In the context of cultural event recognition, we consider temporal information a rather ``asymmetric clue" where time provides an indicator to rather reject a given hypothesis than to support it. On the one hand, given a prediction (e.g. based on visual information) for a photo for a particular event, we can use temporal information, i.e. the capture date of the photo, to easily reject this hypothesis if the capture date does not coincide with the predicted event. In this case temporal information represents a \emph{strong clue}. On the other hand,  cultural events may take place at the same time. As a consequence, the coincidence of a captured date with the predicted event in this case represents just a \emph{weak clue}. We take this ``asymmetric nature" in our temporal refinement scheme (see Section \ref{subsec:tempRefinement}) into account.

Temporal information has further been exploited for event classification by Mattive et al. ~\cite{mattivi2011exploitation}. The authors define a two-level hierarchy of events and sub-events which are automatically classified based on their visual information described as a Bag of Visual Words. All photos are first classified visually. Next, the authors refine the classification by enforcing temporal coherence in the classification for each event and sub-event which considerably improved the purely visual classification.

A similar approach is applied by Bossard et al.~\cite{bossard2013event}, exploiting temporal information to define events as a sequence of sub-events. The authors exploit the temporal ordering of photos and model events as a series of sub-events by a Hidden Markov Model (HMM) to improve the classification.


A very similar problem to Cultural Event Recognition, namely ``Social Event Classification", was formulated in the MediaEval Social Event Detection benchmark in 2013 ~\cite{SED2013, Petkos14icmr}. 
The provided dataset contained 57,165 images from Instagram together with available contextual metadata (time, location and tags) provided by the API.
The classification task considered a first decision level between \textit{event} and \textit{non-event} and, in the case of \textit{event}, eight semantic classes were defined to be distinguished: \textit{concert, conference, exhibition, fashion, protest, sports, theatre/dance, other}.
The results over all participants showed that the classification performance strongly  benefits from  multimodal processing combining content and contextual information. Pure contextual processing as proposed in \cite{ADMRG-SED2013} and \cite{VIT-SED2013} and yielded the weakest results. 
The remaining participants proposed to add visual analysis to the contextual processing. CERTH-ITI \cite{schinas2012_1} combined pLSA on the 1,000 most frequent tags with a dense sampling of SIFT visual features, which were later coded with VLAD. They observed a complementary role between visual and textual modalities.
Brenner and Izquierdo \cite{Brenner14icmr} combined textual features with the global  GIST visual descriptor, which is capable of capturing the spatial composition of the scene.
The best performance in the Social Event Classification task was achieved by ~\cite{Nguyen-MediaEval2013}. They combine processing of textual photo descriptions with the work from ~\cite{mattivi2011exploitation} for visual processing, based on bag of visual words aggregated in different fashions through events.
Their results showed that visual information is the best option to discriminate between \textit{event / non-event} and that textual information is more reliable to discriminate between different event types.

In terms of benchmarking, a popular strategy is to retrieve additional data to extend the training dataset.
The authors of \cite{riga2014_6}, for example, retrieved images from Flickr to build unigram language models of the requested event types and locations in order to enable a more robust matching with the user-provided query. 
We explored a similar approach in for cultural event recognition. Results, however showed that extending the training set this did not improve results but made them even worse.

\section{Temporal models}
\label{sec:temporal}

Cultural events usually occur at a regular basis and thus have a repetitive nature. For example, ``St. Patrick's day" always takes place on March, 17, ``La Tomatina" is always scheduled for the last week of August, and the ``Carneval of Rio" usually takes place at some time in February and lasts for one week. More complex temporal patterns exist, for example, for cultural events coupled to the lunar calender which changes slightly each year. An example is the ``Maslenitsa" event in Russia is which is scheduled for the eighth week before Eastern Orthodox Easter.

The temporal patterns associated with cultural events are a valuable clue for their recognition. A photo captured, for example, in December will very unlikely (except for erroneous date information) show a celebration of St. Patrick's day. While temporal information alone is not sufficient to assign the correct event (many events may take place concurrently), we hypothesize that temporal information provides strong clues that can improve cultural event recognition. 

To start with temporal processing, first temporal models have to be extracted from the data. Temporal models for cultural events can be either generated manually in advance or extracted automatically from metadata of related media. We propose a fully automatic approach to extract temporal models for cultural events. The input to our approach is a set of capture dates for media items that are related to a given event. Capture dates may be, for example, extracted from social media sites like Flickr or from the metadata embedded in the photos (e.g. EXIF information). In a first step, we extract the day and month of the capture dates and convert them into a number $d$ between 1 and 365, encoding the day in the year when the photo was taken. From these numbers, we compute a temporal distribution $T(d)$ of all available capture dates. Assuming that a cultural event takes place annually, it is straight-forward to model the temporal distribution with a Gaussian model. Gaussian modeling works well when a sufficient number of timestamps exists. For sparse data, however, with a few timestamps only, the distribution is likely to become non-Gaussian and thus model fitting fails in generating accurate models. Additionally, the timestamps of photos are often erroneous (or overwritten by certain applications) yielding strong deviations of the ideal distribution. To take the variability that is present in the data into account, a more flexible model is required. We model the distribution $t(d)$ by a piecewise cubic smoothing spline \cite{deboor}. To generate the final model $T$, we evaluate the spline over the entire temporal domain and normalize it between 0 and 1. Given a photo $i$ with a certain timestamp $d_i$, the fitted temporal model $T_c(d_i)$ provides a score $s_c$ that the photo refers to the associated event $c$. The flexible spline model enables the modeling of sparse and non-Gaussian distributions and further to model events with more complex than annual occurrence patterns. 

\begin{figure}%
		\subfigure[Maslenitsa]{\label{sfig:modelExample1}
				\includegraphics[width=0.48\linewidth]{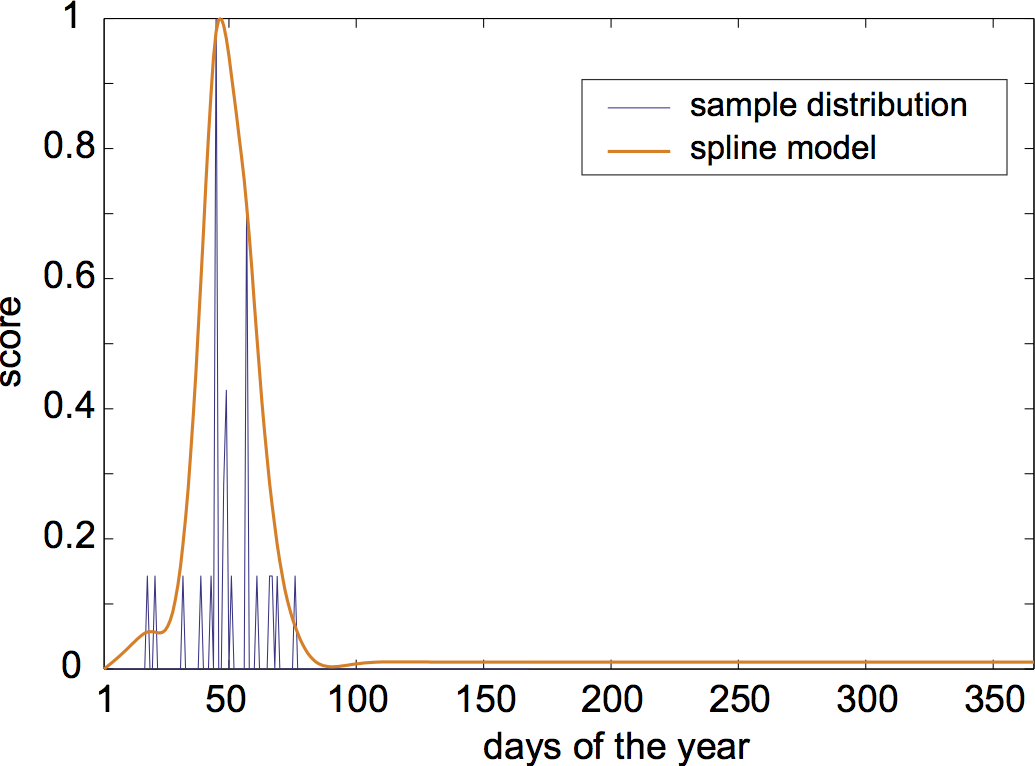}}
		\subfigure[Timkat]{\label{sfig:modelExample2}
				\includegraphics[width=0.48\linewidth]{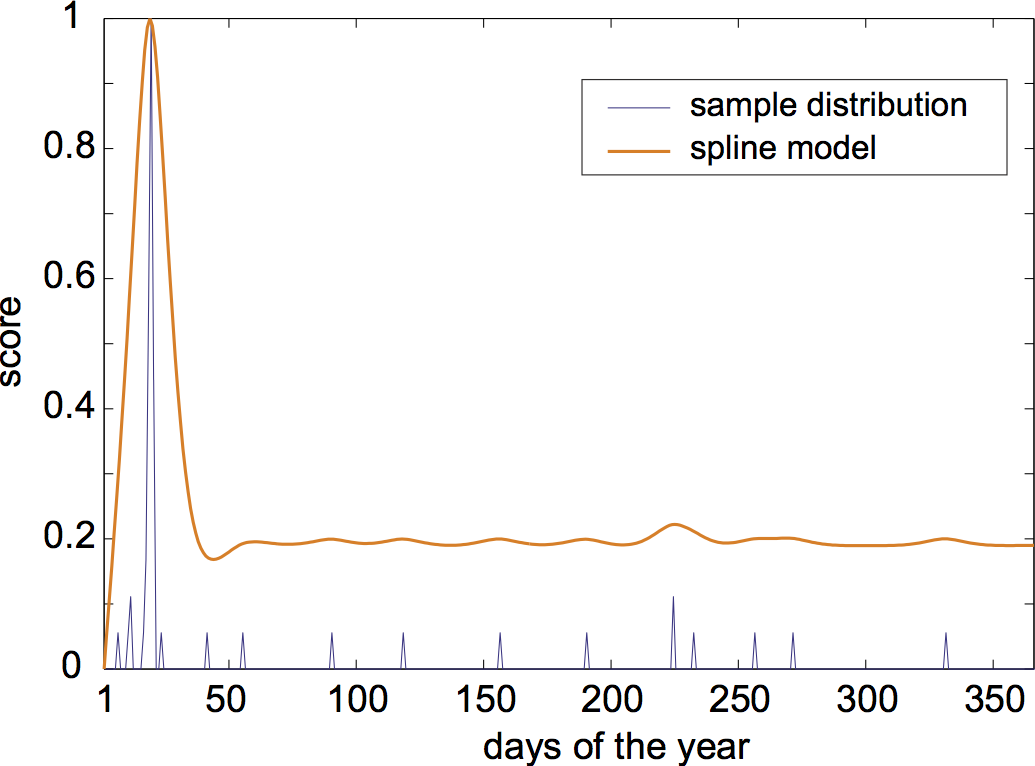}}
		\caption{Temporal spline models for the ``Maslenitsa" and the ``Timkat" event: (a) for normally distributed data the model becomes approximately Gaussian-shaped; (b) the uncertainty of the distribution is reflected in the temporal model.}
		\label{fig:tempModelExamples}
\end{figure}

Figure \ref{fig:tempModelExamples} shows temporal models for two example events. The ``Maslenitsa" (\ref{sfig:modelExample1}) takes place between mid of February and mid of March (approx. days 46-74). This corresponds well with the timestamps extracted from the related media items, resulting in a near Gaussian-shaped model. The ``Timkat" event always takes place on January 19. This is accurately detected by the model, which has its peak at day 19. The photos related to this event, however, have timestamps that are distributed across the entire year. This property of the underlying data is reflected in the model, giving low but non-zero scores to photos with timestamps other than the actual event date.


\begin{figure}%
		\includegraphics[width=0.98\linewidth]{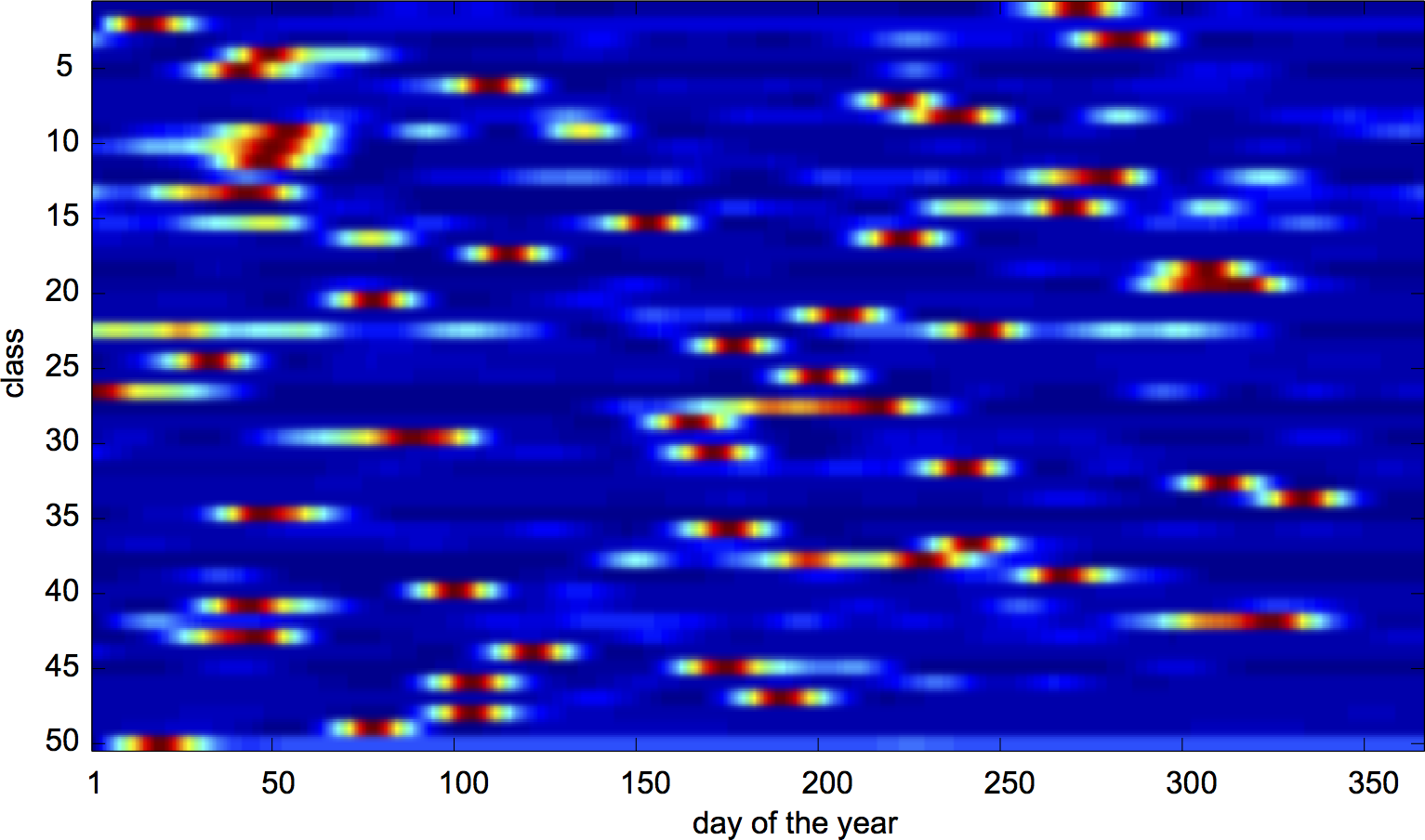}
		\caption{Automatically generated temporal models for each event class. For each event we observe a typical pattern of recording dates exhibiting one strong peak. The colors range from dark blue (0) to red (1).}
		\label{fig:tempModels}
\end{figure}

Figure \ref{fig:tempModels} shows the temporal models extracted from the training and validation data for all 50 classes. We observe that each model (row) exhibits one strong peak which represents the most likely date of the event. Some models contain additional smaller side-peaks learned from the training data which reflect the uncertainty contained in the training data. The events are distributed over the entire year, some events occur at the same time.

The generated temporal models can be used to refine decisions made during classification (see Section \ref{subsec:tempRefinement}) as well as for the filtering of additional data collections to reduce noise in the training data (see Section \ref{subsec:externalDataDownload}).

\section{Image Classification}
\label{sec:classification}

The automatic recognition of a cultural event from a photo is addressed in this paper with the system architecture presented in Figure \ref{fig:pipeline}.
We propose combining the visual features obtained at the fully connected layers of two versions of the same \textit{Caffenet} convolutional neural network: the original one and a modified version fine-tuned with photos captured at cultural events.
A low-level SVM classifier is trained for each visual feature, and its scores refined with the temporal model described in Section \ref{sec:temporal}.
Finally, the temporally modified classification scores are fused in a final high-level SVM to obtain the final classification for a given test image.

\subsection{Feature extraction}
\label{ssec:FeatureExtraction}

Deep convolutional neural networks (convnets) have recently become popular in computer vision, since they have dramatically advanced the state-of-the-art in tasks such as image classification \cite{krizhevsky2012imagenet}, retrieval \cite{babenko2014neural} or object detection \cite{girshick2014rich,hariharan2014simultaneous} 

Convnets are typically defined as a hierarchical structure of a repetitive pattern of three hidden layers: (a) a local convolutional filtering (bidimensional in the case of images), (b) a non-linear operation, (commonly Rectified Linear Units - ReLU) and (c) a spatial local pooling (typically a \textit{max} operator). 
The resulting data structure is called a \textit{feature map} and, in the case of images, they correspond to 2D signals.
The deepest layers in the convnet do not follow this pattern anymore but consist of \textit{fully connected (FC)} layers: every value (neuron) in the fully connected layer is connected to all neurons from the previous layers through some weights.
As these fully connected layers do not apply any spatial constrain anymore, they are represented as single dimensional vectors, further referred in this paper as \textit{neural codes} \cite{babenko2014neural}.


The amount of layers is a design parameter that, in the literature, may vary from three \cite{lecun1998gradient} to nineteen \cite{simonyan2014very}.
Some studies \cite{zeiler2014visualizing} indicate that the first layers capture finer patterns, while the deeper the level, the more complex patterns are modeled.
However, there is no clear answer yet about how to find the optimal architecture to solve a particular visual recognition problem.
The design of convnets is still mainly based on trial-and-error process and the expertise of the designer.
In our work we have adopted the public implementation of \textit{CaffeNet} \cite{jia2014caffe}, which was inspired by \textit{AlexNet} \cite{krizhevsky2012imagenet}.
This convnet is defined by 8 layers, being the last 3 of them fully connected.
In our work we have considered the neural codes in these layers (FC6, FC7 and FC8) to visually represent the image content.

Apart from defining a convnet architecture, it is necessary to learn the parameters that govern the behaviour of the filters in each layer.
These parameters are obtained through a learning process that replaces the classic handcrafted design of visual features.
This way, the visual features are optimized for the specific problems that one wants to solve.
Training a convnet is achieved through backpropagation, a high-computational effort that has been recently boosted by the affordable costs of GPUs.
In addition to the computational requirements, a large amount of annotated data is also necessary.
Similarly to the strategy adopted in the design of the convnet, we have also used the publicly available filter parameters of \textit{CaffeNet} \cite{jia2014caffe}, which had been trained for 1,000 semantic classes from the ImageNet dataset \cite{deng2009imagenet}.

The cultural event recognition dataset aimed in this paper is different from the one used to train \textit{CaffeNet}, both in the type of images and in the classification labels.
In addition, the amount of photos of annotated cultural events available in this work is much smaller than the large amount of images available in ImageNet.
We have addressed the situation by also considering the possibility of fine tuning \textit{CaffeNet}, that is, providing additional training data to an existing convnet which had been trained for a similar problem.
This way, the network parameters are not randomly initialized, as in a training from scratch, but are already adjusted to a solution which is assumed to be similar to the desired one. 
Previous works \cite{girshick2014rich,hariharan2014simultaneous, Chatfield14} have proved that fine-tuning \cite{hinton2006reducing} is an efficient and valid solution to address these type of situations. 
In the experiments reported in Section \ref{sec:experiments} we have used feature vectors from both the original \textit{CaffeNet} and its fine-tuned version.

\subsection{Hierarchical fusion}
\label{ssec:HierarchicalFusion}

The classification approach applied in our work is using the neural codes extracted from the convnets as features to train an classifier (Support Vector Machines, SVMs, in our case), as proposed in \cite{Chatfield14}. As we do not know a priori which network layer are most suitable for our task, we decide to combine several layers using a late fusion strategy.

The neural codes obtained from different networks and different layers may have strongly different dimensionality (e.g. from 4,096 to 50 in our setup). During the fusion of these features we have to take care that features with higher dimensionality do not dominate the features with lower dimensionality. Thus, we adopted a hierarchical classification scheme to late fuse the information from the different features in a balanced way \cite{snoek2005early}.

At the lower level of the hierarchy we train separate multi-class SVMs (using one-against-one strategy \cite{hsu2002comparison}) for each type of neural code. We neglect the final predictions of the SVM and retrieve the probabilities of each sample for each class. The probabilities obtained by all lower-level SVMs form the input to the higher hierarchy level. 

The higher hierarchy level consists of an SVM that takes to probabilistic output of the lower-level SVMs as input. This assures that all input features are weighted equally in the final decision step. The higher-level SVM is trained directly from the probabilities and outputs a prediction for the most likely event. Again we reject the binary prediction and retrieve the probabilities for each event as the final output.

\subsection{Temporal Refinement}
\label{subsec:tempRefinement} 

While visual features can easily be extracted from each image, the availability of temporal information depends on the existence of suitable metadata. Thus, temporal information must in general be considered to be a sparsely available feature. Due to its sparse nature, we propose to integrate temporal information into the classification process by refining the classifier outputs. This allows us to selectively incorporate the information only for those images where temporal information is available.

The basis for temporal refinement are the temporal models introduced in Section \ref{sec:temporal}. The models $T_c$ with $c=1,\ldots,C$ and $C$ the number of classes, represent for each event class $c$ and each day of the year $d$, a score $s$ representing the probability of a photo captured in a given day to belong to the event: $s=T_c(d)$. For a given image with index $i$, we first extract the day of the year $d_i$ from its capture date and use it as an index to retrieve the scores from the temporal models of all event classes: $s_c = T_c(d_i)$, with $s=\left\{s_1,\ldots,s_C\right\}$.

Given a set of probabilities $P_i$ for image $i$ obtained from a classifier, the refinement of these probabilities is performed as follows. First, we compute the difference between the probabilities and the temporal scores: $d_i = P_i - s$. Next, we distinguish between two different cases:

\textbf{(I) $d_i(c)<0$}: Negative differences mean that the probability for a given class predicted by the classifier is less than the temporal score for this class. This case may easily happen as several events may occur at the same time as the photo was taken. The temporal models indicate that several events may be likely. Thus, the temporal information provides only a weak clue that is not discriminative. To handle this case, we decide to trust the class probabilities by the classifier and to ignore the temporal scores by setting $d=max(d,0)$. 

\textbf{(II) $d_i(c)>0$}. In this case the temporal score is lower than the estimate of the classifier. Here, the temporal score provides a strong clue that indicates an inaccurate prediction of the classifier. In this case, we use the difference $d_i(c)$ to re-weight the class probability.

The weights $w_i$ are defined as $w_i=max(d,0)+1$. The final re-weighting of the probabilities $P_i$ is performed by computing $\tilde{P}_i=P_i/w_i$. In case \textbf{(I)} the temporal scores do not change the original predictions of the classifier. In case \textbf{(II)} the scores are penalized by a fraction that is proportional to the disagreement between the temporal scores and the prediction of the classifier.


\section{Data Augmentation}
\label{sec:DataAugmentation}

The experiments described in Section \ref{sec:experiments} were conducted with the ChaLearn Cultural Event Recognition dataset \cite{escalera2015chalearn}, which was created by downloading photos from \textit{Google Images} and \textit{Bing} search engines.
Previous works \cite{krizhevsky2012imagenet, zeiler2014visualizing, Chatfield14} have reported gains when applying some sort of data augmentation strategy.

We have explored two paths for data augmentation: artificial transformations on the test images and an extension of the training dataset by downloading additional data from Flickr.

\subsection{Image transformations}

A simple and classic method for data augmentation is to artificially generate transformations of the test image and fuse the classification scores obtained in each transformation.
We adopted the default image transformations associated to \textit{CaffeNet} \cite{jia2014caffe}, this is an horizontal mirroring and 5 crops in the input image (four corners and center).
The resulting neural codes associated to each fully connected layer were fused by averaging the 10 feature vectors generated with the 10 image transformations.

\subsection{External data download}
\label{subsec:externalDataDownload}

We decided to extend the amount of training data to fine-tune our convnet, as discussed in Section \ref{ssec:FeatureExtraction}.
By doing this, we expected to reduce the generalization error of the learned model by having examples coming from a wider origin of sources.

The creators of the ChaLearn Cultural Event Recognition dataset \cite{escalera2015chalearn} described each of the 50 considered events with pairs of title and geographical location; such as \textit{Carnival Rio-Brazil}, \textit{Obon-Japan} or \textit{Harbin Ice and Snow Festival-China}.
This information allows generating queries on other databases to obtained an additional set of labeled data.

Our chosen source for the augmented data was the \textit{Flickr} photo repository. 
Its public API allows to query its large database of photos and filter the obtained results by tags, textual data search and geographical location.
We generated 3 sets of images from Flickr, each of them introducing a higher degree of refinement:

\begin{description}
\item[90k set:] Around 90,000 photos retrieved by matching the provided event title on the Flickr \textit{tags} and \textit{content} metadata fields.
\item[21k set:] The query from the 90k set was combined with a GPS filtering based on the provided country.
\item[9k set:] The query from the 21k set was further with manually selected terms from the Wikipedia articles related to the event.
In addition, the Flickr query also toggled on an \textit{interestingness} flag which improved the diversity of images in terms of users and dates.
Otherwise, Flickr would provide a list sorted by upload date, which will probably contain many similar images from a reduced set of users.
\end{description}



\begin{figure}%
	\subfigure[Desfile de Silleteros]{\label{sfig:flickrExample1}
		\includegraphics[width=0.9\linewidth]{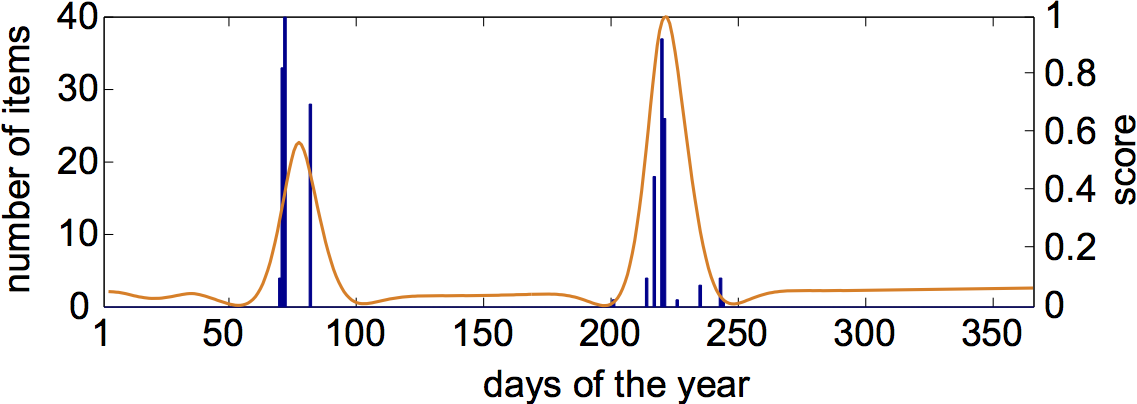}}
				
	\subfigure[Carnival of Venice]{\label{sfig:flickrExample2}
		\includegraphics[width=0.9\linewidth]{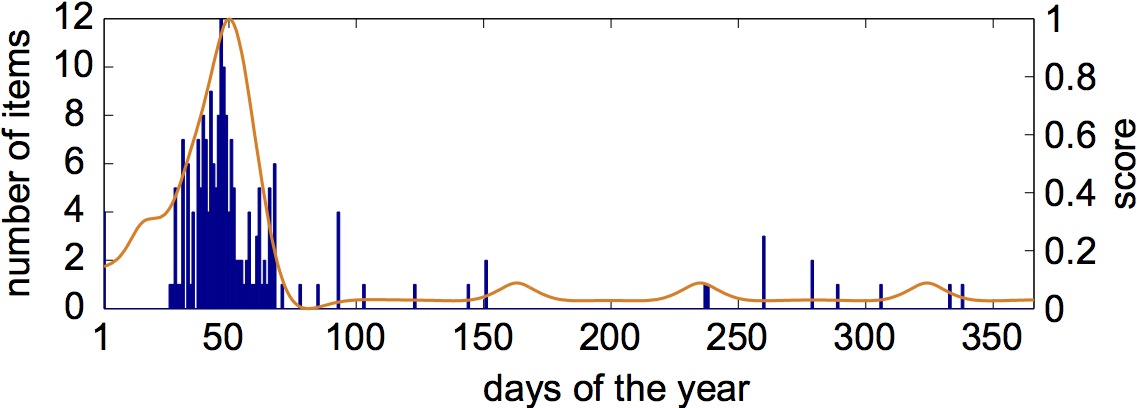}}
	\caption{Two examples of retrieved image collections from Flickr and their temporal distribution. (a) the retrieved images match well the pre-trained temporal model. (b) the temporal distribution shows numerous outliers which are considered unlikely given the temporal model. The proposed threshold-based filtering removes those items.}
	\label{fig:flickrFilteringExamples}
\end{figure}

The temporal models $T_c$ presented in Section \ref{sec:temporal} were also 
used to improve the likelihood that a downloaded photo actually belongs to a certain event.
Given a media item $i$ retrieved for a given event class $c$, we extract the day of capture $d_i$ from its metadata and retrieve the score $s_c=T_c(d_i)$ from the respective temporal model. Next, we threshold the score to remove items that are unlikely under the temporal model. To assure a high precision of the filtered media collection, the threshold should be set to a rather high value, e.g. 0.9. Figure \ref{fig:flickrFilteringExamples} gives two examples of media collections retrieved for particular events. We provide the distribution of capture dates with the pre-trained temporal models.

The Flickr IDs of this augmented dataset filtered by minimum temporal scores have been published in JSON format from the URL indicated in Section \ref{sec:motivation}.

\section{Experiments}
\label{sec:experiments}

\subsection{Cultural Event Recognition dataset}

The Cultural Event Recognition dataset \cite{escalera2015chalearn} 
depicts 50 important cultural events all over the world. 
In all the image categories, garments, human poses, objects and context do constitute the possible cues to be exploited for recognizing the events, while preserving the inherent inter- and intra-class variability of this type of images. The dataset is divided in three partitions: 5,875 images for \textit{training}, 2,332 for \textit{validation} and 3,569 for \textit{test}.

\subsection{Experimental setup}

We employ two different convnets as input (see Section \ref{ssec:FeatureExtraction}): the original \textit{CaffeNet} trained on 1,000 Imagenet classes, and a fine-tuned version of \textit{CaffeNet} trained during 60 epochs on the 50 classes defined in the Chalearn Cultural Recognition Dataset.
Fine-tuning of the convnet was performed in two stages: in a first one the \textit{training} partition was used to train and the \textit{validation} partition to estimate the training loss and allow the network to learn. 
In a second stage, the two partitions were switched so that the network had to learn the optimal features from all the available labeled data.

From both convnets we extracted neural codes from layers FC6 and FC7 (each of 4,096 dimensions), as well as FC8 (the top layer with a softmax classifier), which has 1000 dimensions for the original \textit{CaffeNet} and 50 for the fine-tuned network.
Both feature extraction and fine tuning have been performed using the \emph{Caffe} \cite{jia2014caffe} deep learning framework.


As presented in Section \ref{ssec:HierarchicalFusion}, a classifier was trained for each of the 6 neural codes, in addition to the one used for late fusion.
The implementation of \textit{Libsvm} library \cite{chang2011libsvm}  of the linear SVM was used, with parameter $C=1$ determined by cross validation and grid search and probabilistic output switched on.
%
%
%
%
%

Each image was scored for each of the 50 considered cultural events and results were measured by a precision/recall curve, whose area under the curve was used to estimate the average precision (AP).
Numerical results are averaged over the 50 events to obtain the mean average precision (mAP).
More details about the evaluation process can be found in \cite{escalera2015chalearn}.


 

\subsection{Results on the validation dataset}

A first experimentation was performed to assess the impact of temporal refinement on the default \textit{CaffeNet}, that is, with no fine-tunning.
Results in Table \ref{tab:SingleLayer} indicate diverse performance among the fully connected layers, being FC6 the one with a highest score. Temporal refinement slightly increases the mAP consistently in all layers. 

\begin{table}
\begin{center}
\begin{tabular}{|l|c|c|c|}
\hline
							&	FC6 	& FC7 		& FC8 \\
\hline\hline
Raw layer				 & 0,6832 	& 0,6669	& 0,6079 \\
+ temporal refinement 	& 0,6893 	& 0,6730	& 0,6152 \\
\hline
\end{tabular}
\end{center}
\caption{Results on single layer raw neural codes.}
\label{tab:SingleLayer}
\end{table}

The preliminary results were further extended to compare the performance of the three neural codes (FC6, FC7 and FC8) when temporally refined and finally complemented with the features from the original \textit{CaffeNet}.
The results shown in Table \ref{tab:Multilayer} indicate a higher impact of temporal refinement than in the case of single layers, and an unexpected gain by adding the raw neural codes from \textit{CaffeNet}.

\begin{table}
\begin{center}
\begin{tabular}{|l|c|}

\hline
Fine-tunned FC6-FC7-FC8		& 0,6919 	\\
+ raw FC6-FC7-FC8			& 0,7038	\\
+ temporal refinement		& 0,7357	\\
\hline
\end{tabular}
\end{center}
\caption{Results on fine-tuned and fused multi-layer codes.}
\label{tab:Multilayer}
\end{table}

Our experimentation on the additional data downloaded from Flickr was unsuccessful. The selected dataset was the 9k Flickr one with a restrictive threshold of $0.9$ on the temporal score. With this procedure we selected 5,492 images, which were added as training samples for fine tuning.
%
%
%
%
We compare the impact of adding this data into training only on the softmax classifier at the last layer of \textit{CaffeNet}, obtaining a drop in the mAP from $0.5821$ to $0.4547$ when adding the additional images to the already fine-tuned network.
We hypothesize that the visual nature of the images downloaded from Flickr differs from the one of the data crawled from Google and Bing by the creators of the ChaLearn dataset.
A visual inspection on the augmented dataset did not provide any hints that could expalin this behaviour.

\subsection{Results on the test dataset}

The best configuration obtained with the validation dataset was used on the test dataset to participate in the ChaLearn 2015 challenge.
Our submission was scored by the organizers with a mAP of $0,767$, the second best performance among the seven teams which completed the submission, out of the 42 participants who had initially registered on the challenge website.

\section{Conclusions}
\label{sec:conclusions}

The presented work proves the high potential of the visual information for cultural event recognition.
This result is especially sounding when contrasted with many of the conclusions made in the MediaEval Social Event Detection task \cite{Petkos14icmr}, where it was frequently observed that visual information was less reliable than contextual metadata for event clustering.
This difference may be caused by the very salient and distinctive visual features that often make cultural events attractive and unique.
The dominant green in Saint Patrick's parades, the vivid colors from the Holi Festivals or the skull icons from the Dia de los Muertos 

In our experimentation the temporal refinement has provided modest gain. 
We think this may be caused by the low portion of images with available EXIF metadata, around $24\%$ according to our estimations.
In addition, we were also surprised by the loss introduced by the Flickr data augmentations.
We plan to look at this problem more closely and figure out the difference between the ChaLearn dataset and ours.

Finally, it must be noticed that the quantitative values around $0.7$ may be misleading, as in this dataset every image belonged to one of the 50 cultural events.
Further editions of the ChaLearn challenge may also introduce the \textit{no event} class as in MediaEval SED 2013 \cite{SED2013} to, this way, better reproduce a  realistic scenario where the event retrieval is performed in the wild.






\section*{Acknowledgements}
\label{sec:acnkowledgements}

This work has been developed in the framework of the project BigGraph TEC2013-43935-R, funded by the Spanish Ministerio de Economía y Competitividad and the European Regional Development Fund (ERDF). 

The Image Processing Group at the UPC is a SGR14 Consolidated Research Group recognized and sponsored by the Catalan Government (Generalitat de Catalunya) through its  AGAUR office.

We gratefully acknowledge the support of NVIDIA Corporation with the donation of the GeoForce GTX Titan Z used in this work.

{\small
\bibliographystyle{ieee}
\bibliography{main_final}
}

\end{document}